\documentclass[10pt,twocolumn,letterpaper,capitalize]{article}

\usepackage[pagenumbers]{style-cvpr/cvpr}      

\usepackage[utf8]{inputenc} 
\usepackage[T1]{fontenc}    
\usepackage{url}            
\usepackage{nicefrac}       
\usepackage{microtype}      
\usepackage[usenames, dvipsnames]{xcolor}         
\usepackage{comment}

\usepackage{booktabs}   
\usepackage{tabularx}   
\usepackage{colortbl}   
\usepackage{multirow}
\usepackage{makecell}
\renewcommand{\arraystretch}{1.1}

\usepackage{wrapfig}
\usepackage{mwe} 

\usepackage[font=normal,labelfont=bf]{caption}
\usepackage[font=normal]{subcaption}

\usepackage[normalem]{ulem} 
\usepackage{array}

\usepackage{xparse}
\usepackage{pifont}

\usepackage{ifthen}

\definecolor{cvprblue}{rgb}{0.21,0.49,0.74}
\usepackage[pagebackref,breaklinks,colorlinks,citecolor=cvprblue]{hyperref}

\usepackage[capitalize]{cleveref}
\crefformat{section}{Section~#2#1#3}
\crefformat{subsection}{Section~#2#1#3}
\crefformat{subsubsection}{Section~#2#1#3}
\crefformat{figure}{Fig.~#2#1#3}
\crefformat{equation}{Eq.~#2#1#3}
\crefformat{table}{Table~#2#1#3}
\crefformat{algorithm}{Alg.~#2#1#3}
\usepackage{algpseudocode}
\usepackage[linesnumbered,ruled,vlined]{algorithm2e}

\SetCommentSty{mycommfont}
\SetKwInput{KwInput}{Input}

\usepackage{enumitem}

\usepackage{xfp}

\newcommand{\method}{{\fontfamily{lmtt}\selectfont\emph{\textsc{DeepAVFusion}}}\xspace}

\newcommand{\cmark}{\ding{51}}%
\newcommand{\xmark}{\ding{55}}%

\newlength\savewidth
\newlength\thinwidth

\definecolor{Gray}{gray}{0.93}
\newcolumntype{a}{>{\columncolor{Gray}}c}

\usepackage{xspace} 
\makeatletter
\DeclareRobustCommand\onedot{\futurelet\@let@token\@onedot}
\def\@onedot{\ifx\@let@token.\else.\null\fi\xspace}
\def\eg{\emph{e.g}\onedot} 
\def\ie{\emph{i.e}\onedot}

\makeatother

\newcommand{\circled}[1]{\raisebox{.5pt}{\textcircled{\raisebox{-.9pt} {#1}}}}

\definecolor{battleshipgrey}{rgb}{0.52, 0.52, 0.51}

\usepackage{amsmath,amsfonts,bm}

\def\1{\bm{1}}

\def\vis{{\text{vis}}}
\def\mask{{\text{mask}}}

\def\mK{{\bm{K}}}

\def\mM{{\bm{M}}}

\def\mQ{{\bm{Q}}}

\def\mV{{\bm{V}}}

\def\mX{{\bm{X}}}

\def\mZ{{\bm{Z}}}

\DeclareMathAlphabet{\mathsfit}{\encodingdefault}{\sfdefault}{m}{sl}
\SetMathAlphabet{\mathsfit}{bold}{\encodingdefault}{\sfdefault}{bx}{n}

\def\gL{{\mathcal{L}}}
\def\gM{{\mathcal{M}}}

\def\gV{{\mathcal{V}}}

\setlist{itemsep=0.5ex,parsep=0.5ex,leftmargin=0.4cm}
\setenumerate{itemsep=1ex,parsep=0.3ex}
\setitemize{itemsep=0ex,parsep=0.3ex,leftmargin=0.4cm}

\renewcommand{\paragraph}[1]{\noindent {\bf #1}}

\def\paperID{} 
\def\confName{CVPR}
\def\confYear{2024}

\title{Unveiling the Power of Audio-Visual Early Fusion Transformers with Dense Interactions through Masked Modeling}

\author{%
  Shentong Mo\thanks{Corresponding author.}\\
  Carnegie Mellon University
  \and
  Pedro Morgado\\
  University of Wisconsin-Madison 
  \and 
  \url{https://github.com/stoneMo/DeepAVFusion}
}

\begin{document}

\maketitle

\begin{abstract}

Humans possess a remarkable ability to integrate auditory and visual information, enabling a deeper understanding of the surrounding environment. This early fusion of audio and visual cues, demonstrated through cognitive psychology and neuroscience research, offers promising potential for developing multimodal perception models. However, training early fusion architectures poses significant challenges, as the increased model expressivity requires robust learning frameworks to harness their enhanced capabilities. In this paper, we address this challenge by leveraging the masked reconstruction framework, previously successful in unimodal settings, to train audio-visual encoders with early fusion. Additionally, we propose an attention-based fusion module that captures interactions between local audio and visual representations, enhancing the model's ability to capture fine-grained interactions. While effective, this procedure can become computationally intractable, as the number of local representations increases. Thus, to address the computational complexity, we propose an alternative procedure that factorizes the local representations before representing audio-visual interactions. Extensive evaluations on a variety of datasets demonstrate the superiority of our approach in audio-event classification, visual sound localization, sound separation, and audio-visual segmentation. These contributions enable the efficient training of deeply integrated audio-visual models and significantly advance the usefulness of early fusion architectures.


\end{abstract}

\section{Introduction}
Humans naturally integrate audio-visual information to perceive and understand the environment. Several studies in cognitive psychology and neuroscience, as well as classic examples such as the McGurk effect, demonstrate that such audio-visual fusion can occur early on in the perceptual stack, enabling deeper integration of the two modalities.
Early fusion models aim to emulate this human-like perception. Specifically, it refers to the process of integrating auditory and visual cues \textit{at an early stage} of a multi-modal perception model, in order to leverage the synergistic effects of both modalities. 
This can be especially powerful if the fusion process can attend to and establish connections between local components of the visual and audio signal (\eg, the frequencies of someone's voice with the pixels of their lips).
In sum, early fusion of local audio-visual interactions holds the promise of a deeper and more sophisticated understanding of audio-visual content, critical for many real-world applications such as visually guided source separation, localization/segmentation, and multi-modal recognition.

Despite their potential, audio-visual early fusion architectures have remained under-explored, with the majority of the recent literature focusing on late fusion or no fusion at all.
For example, one prominent research direction~\cite{Arandjelovic2017look,Morgado2021audio} seeks to learn independent uni-modal encoders through contrastive learning by requiring the representations of associated audio-visual pairs to be synchronized in latent space. However, contrastive learning techniques are not compatible with early fusion, as the connections between uni-modal representations lead to trivial solutions to the contrastive learning problem.
Another line of recent work~\cite{huang2022mavil,gong2023contrastive} learns encoders with late audio-visual fusion through a combination of contrastive learning and masked reconstruction, focusing on downstream tasks that require shallow integration of audio-visual features, such as audio-visual action recognition.

\begin{figure}[t]
\centering
\includegraphics[width=\linewidth]{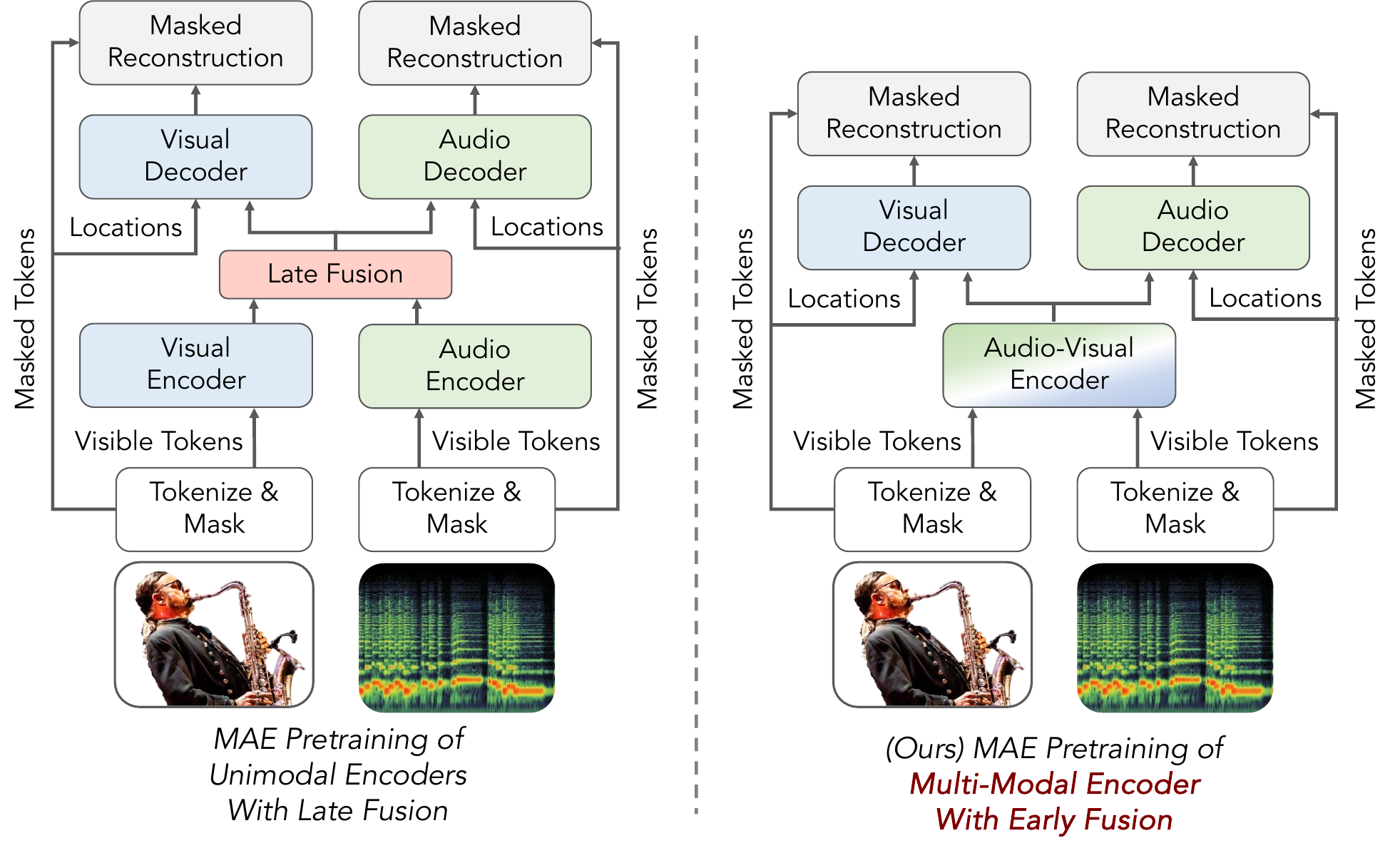}
\caption{Unlike prior works on Audio-Visual MAEs which learn either late fusion encoders or separate encoders altogether, we demonstrate that masked reconstruction is especially useful for learning deeply integrated audio-visual encoders with powerful early fusion modules.}
\label{fig: teaser}
\end{figure}

One potential reason is that early fusion architectures pose a significant training challenge due to the substantial increase in model expressivity it offers. As a result, robust learning frameworks are required to ensure the creation of high-quality models and effectively harness their enhanced capabilities.
In this work, we demonstrate the effectiveness of a masked reconstruction framework for training audio-visual transformers with early fusion of local interactions (\cref{fig: teaser}). Our hypothesis is that the simultaneous reconstruction of audio and visual inputs from a limited audio-visual context promotes the learning of deeply integrated representations, as the model is required to understand the fine interactions between the two modalities.
Furthermore, masked reconstruction has already been successfully deployed in training uni-modal representations for a variety of input signals, including text~\cite{devlin2019bert}, images~\cite{bao2022beit,he2021masked}, and audio~\cite{huang2022amae}, as well as for learning late fusion representations~\cite{huang2022mavil,gong2023contrastive}. 
However, unlike what was previously observed with uni-modal or late fusion models, our work shows that masked audio-visual modeling with early fusion encoders results in an interesting emergent property. 
While uni-modal representations encode the low-level details required for masked reconstruction, we observed that high-level semantics (surprisingly) emerged from the tokens used for audio-visual fusion.

In addition to demonstrating the effectiveness of masked reconstruction for early fusion transformers, we propose a novel attention-based fusion module that can effectively attend to \textit{interactions between local audio and visual representations}, thus enhancing its ability to capture localized interactions between the two modalities. 
For example, the sound of a dog barking indicates not only the presence of a dog but also provides cues regarding its state (\eg open mouth). While the association between the dog and barking can be established based on global representations, the proposed fusion module enables the model to further capture local interactions (\eg between the dog's mouth and the spectral frequencies associated with barking).
Although localized fusion enhances the learned representations for a variety of tasks, this requires the model to explicitly represent the interactions between all visual and audio tokens. However, given the large number of tokens processed by current transformers, dense local interactions can become easily intractable. To further improve its efficiency, the proposed fusion module factorizes uni-modal representations to avoid exhaustively modeling all pairwise interactions.

In sum, we proposed an attention-based fusion module that can attend to local audio-visual interactions, and deployed it to learn deeply integrated audio-visual representations by fusing uni-modal representations at early layers in the architecture. We also demonstrate that when paired with masked reconstruction, our framework, denoted \method, can learn strong audio-visual representations, yielding state-of-the-art performance on a variety of audio-visual tasks, including visually guided source separation, localization, and segmentation.
We extensively evaluated \method on a variety of audio-visual datasets, and conducted thorough ablation studies, showing the importance of early fusion, dense interactions, and uni-modal pre-training when learning deeply integrated audio-visual representations.

\section{Related Work}
\vspace{-0.7em}

\noindent{\textbf{Audio-Visual Representations Learning.}}
Audio-visual representations learning has been addressed in many previous works~\cite{aytar2016soundnet,owens2016ambient,Arandjelovic2017look,korbar2018cooperative,Senocak2018learning,zhao2018the,zhao2019the,Gan2020music,Morgado2021robust,Morgado2021audio,hershey2001audio,ephrat2018looking,hu2019deep,mo2022semantic,mo2023diffava,pian2023audiovisual,mo2023class} to learn the audio-visual correlation between two distinct modalities from videos.
Such cross-modal alignments are beneficial for many audio-visual tasks, such as audio-event localization~\cite{tian2018ave,lin2019dual,wu2019dual,lin2020audiovisual,mo2022EZVSL,mo2022benchmarking,mo2023audiovisual,mo2023avsam,mo2023weaklysupervised}, audio-visual spatialization~\cite{Morgado2018selfsupervised,gao20192.5D,Chen2020SoundSpacesAN,Morgado2020learning}, audio-visual navigation~\cite{Chen2020SoundSpacesAN,chen2021waypoints,chen22soundspaces2} and audio-visual parsing~\cite{tian2020avvp,wu2021explore,lin2021exploring,mo2022multimodal}.
In this work, our main focus is to learn transferrable audio-visual representations from masked audio-visual reconstruction, which is more challenging than the above-mentioned tasks.

\noindent{\textbf{Masked Representations Learning.}}
Masked representation learning aims to learn self-supervised representations by reconstructing desired features of masked data given unmasked parts as clues.
In the recent years, masked representation learning has achieved promising results in natural language processing~\cite{devlin2019bert,liu2019roberta,sun2019ernie,Conneau2019xlm,wettig2023mask} and computer vision~\cite{bao2022beit,he2021masked,wei2022masked,xie2022SimMIM,chen2022cae,wu2022objectwise,feichtenhofer2022masked,dong2023peco} community.
Typically, BERT~\cite{devlin2019bert} randomly masked 15\% of word tokens in the sentence and recovered them with unmasked words to learn generalizable textual features via a self-attention transformer~\cite{ashish2017attention}.
A block-wise masking strategy was proposed in BEiT~\cite{bao2022beit} to reconstruct discrete tokens of masked image patches for pre-training transferrable visual representations.
To simplify the masked image encoding framework, MAE~\cite{he2021masked} directly reconstructed missing pixels of 75\% masked patches using vision transformers~\cite{Dosovitskiy2021vit} for self-supervised pre-training.

\noindent{\textbf{Audio-visual Masked Autoencoders.}}
More recently, researchers introduced diverse masking pipelines~\cite{huang2022mavil,gong2023contrastive,Georgescu2023audiovisual} to show the effectiveness of masked modeling in learning audio-visual representations.
For example, CAV-MAE~\cite{gong2023contrastive} combined contrastive learning and masked modeling to capture a joint and coordinated audio-visual representation.
They tried to add a joint encoder on audio-visual features from the last attention block of single-modality encoders to fuse cross-modal information for audio-visual contrastive objectives. 
MAViL~\cite{huang2022mavil} extended masked audio-video reconstruction with masked intra- and inter-modal contrastive learning and self-training by recovering joint audio-video contextualized representations.
Despite their promising performance, they ignored the importance of the early fusion of audio-visual features in masked audio-visual reconstruction.
In contrast to AV-MAE~\cite{Georgescu2023audiovisual} that simply concatenated the audio and visual tokens before passing them through the joint transformer for early fusion, we will design an early fusion module with interactions between local and visual representations for audio-visual masked auto-encoders.
Our audio-visual interactions are different from fusion bottlenecks in MBT~\cite{nagrani2022attention} that forces information between different modalities to pass through a small number of bottleneck latent.
However, we develop a fully novel attention-based fusion module that can effectively attend to interactions between local audio and visual representations, thus enhancing its ability to capture fine-grained interactions between the two modalities.

\noindent{\textbf{Audio-visual Early Fusion.}}
Audio-visual fusion has been the topic of extensive research, with several works proposing a variety of architectures to aggregate multi-modal representations~\cite{Owens2018audio,nagrani2022attention,Georgescu2023audiovisual}.
An example of an early fusion architecture was proposed by Owens \textit{et al.}~\cite{Owens2018audio} to learn representations from audio-visual correspondences, by concatenating features from small unimodal encoders and feeding them into a fused audio-visual network for the early fusion of multisensory features. 
Beyond this early work, most recent papers on audio-visual fusion~\cite{nagrani2022attention,Georgescu2023audiovisual} propose alternative fusion mechanisms, either through weight sharing~\cite{Georgescu2023audiovisual} or token-based fusion~\cite{nagrani2022attention}. However, since all these works focused on downstream tasks (mostly classification) that do not require fine multi-modal understanding, the benefits of early fusion were not evident, with mid-level fusion often achieving optimal performance. Different from them, we proposed an early fusion model that is designed for fine multi-modal understanding, by attending to local interactions between the audio and the visual while performing early fusion. The new architecture design paired with masked reconstruction objective is shown to outperform many of the prior works on a variety of downstream multi-modal applications beyond classification.

\begin{figure*}[t]
\centering
\includegraphics[width=0.95\linewidth]{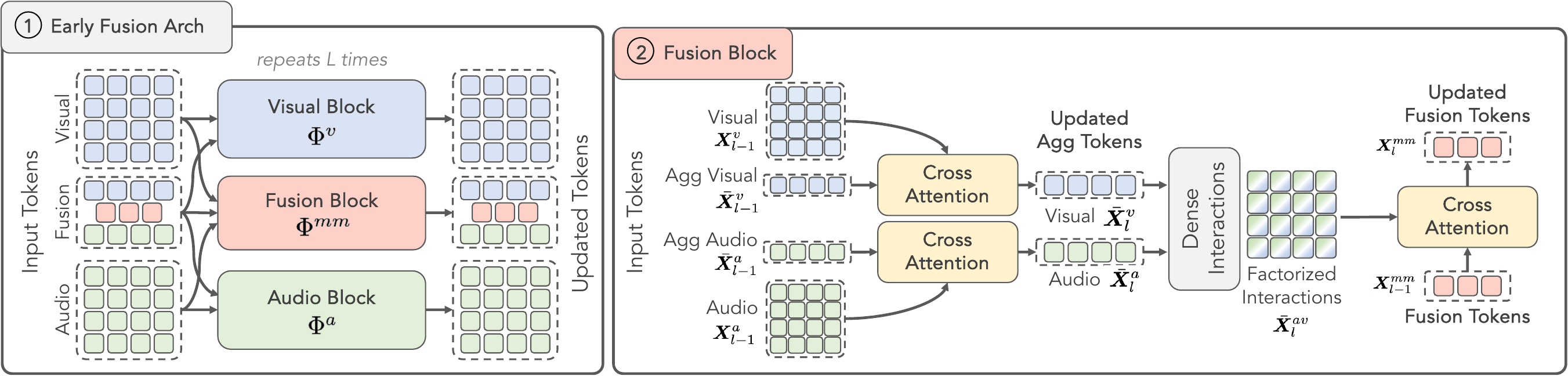}
\caption{Illustration of the proposed early fusion encoder architecture with dense interactions. The architecture \circled{1} is composed of three interconnected branches - the visual, audio, and fusion branches - each with an equal number of transformer blocks. The visual and audio branches process visual and audio tokens, respectively, while simultaneously attending to fusion tokens. The fusion branch updates a set of learnable fusion tokens to fuse audio-visual information. To better attend to local audio-visual interactions, we introduce a new fusion block \circled{2}. At each layer $l$, all patch-level audio (visual) tokens, $\mX^a_{l-1} (\mX^v_{l-1})$, are first aggregated into a small set of aggregation tokens, $\bar{\mX}^a_l (\bar{\mX}^v_l)$ by cross-attention. Pairwise audio-visual interactions are then encoded through a linear layer into latents $\bar{\mX}^{av}_l$, which are used to update the fusion tokens $\mX^{mm}_l$ by cross-attention.}
\label{fig: main_img}
\end{figure*}

\section{Method}
Our goal is to learn audio-visual representations that can be effectively transferred to a variety of downstream tasks that require a detailed understanding of audio-visual interactions. 
To accomplish this, we introduce a novel audio-visual transformer architecture, named \method, that enables the early fusion of audio and visual tokens through dense local interactions. We also show how to effectively train such early fusion transformers using a multi-modal masked reconstruction framework. 
In this section, we first describe the proposed early fusion transformer architecture, and then introduce the multi-modal masked reconstruction framework used for self-supervised pre-training.

\subsection{A Transformer Architecture for Joint Audio-Visual Encoding}
\label{sec:early_fusion}
Downstream tasks such as visually guided sound source separation, localization, and segmentation require a deep understanding of fine audio-visual interactions present in the data. However, late fusion often lacks the expressive power to represent such interactions. We introduce a transformer architecture that enables the early fusion of multi-modal representations through factorized local interactions between audio and visual tokens. 

\paragraph{Early fusion architecture}
While we aim to learn deeply integrated audio-visual representations, foundational uni-modal transformers can still provide a strong starting point for multi-modal learning. As such, we introduce a modular architecture to repurpose existing pre-trained uni-modal transformers. Specifically, we propose a three-branch architecture, illustrated in~\cref{fig: main_img}. 
The first two branches are uni-modal transformers that process audio and visual tokens, respectively, while the third branch is a multi-modal transformer that updates a set of learnable fusion tokens to fuse audio-visual information.
Formally, let $\mX^{mm}_0\in\Re^{F\times D}$ be $F$ learnable fusion tokens. We progressively update these tokens, at each layer $l$, through a fusion block $\Phi^{mm}_l$, which can interact with and aggregate modality-specific features $\mX^v_{l-1}, \mX^a_{l-1}$. Furthermore, fusion tokens $\mX^{mm}_{l-1}$ can be used by the modality-specific blocks, $\Phi^v_l$ and $\Phi^a_l$, to modulate uni-modal representations from early stages in the network. 
In sum, at each layer, representations are updated as follows
\begin{eqnarray}
    \mX^{mm}_l &=& \Phi^{mm}_l(\mX^{mm}_{l-1}, \mX^{v}_{l-1}, \mX^{a}_{l-1})\\
    \mX^{v}_l  &=& \Phi^v_l(\mX^{v}_{l-1}, \mX^{mm}_{l-1})\\
    \mX^{a}_l  &=& \Phi^a_l(\mX^{a}_{l-1}, \mX^{mm}_{l-1})
\end{eqnarray}
This token update strategy is repeated for a total of $L$ layers to compute the final representations $\mX^{v}_{L}, \mX^{a}_{L}$ and $\mX^{mm}_{L}$.

\paragraph{Modality-specific blocks} 
We use standard transformer self-attention for the modality-specific blocks, $\Phi^v$ and $\Phi^a$ and, unless otherwise specified, we initialize these models using foundational uni-modal models. In particular, we empirically validated the benefits of bootstrapping from foundational models using uni-modal MAE pre-trained models~\cite{huang2022amae} and~\cite{he2021masked}.
It should be noted that, in the proposed architecture, modality-specific blocks can also attend to fusion tokens (unlike in their foundational architectures). However, we observed that self-attention could be easily tuned to attend to the additional tokens without requiring new parameters.

\paragraph{Fusion blocks}
The design of the fusion blocks $\Phi^{mm}$ is critical to represent fine audio-visual interactions. 
In the context of transformers, a natural way to combine audio-visual representations is through a standard self-attention block which in addition to the fusion tokens $\mX^{mm}$ themselves, it can further attend to all image and audio tokens, $\mX^{v}_{l-1}$ and $\mX^{a}_{l-1}$. This approach, which we refer to as \textit{token fusion}, has been used in previous works such as MBT~\cite{nagrani2022attention}.
While intuitive, standard self-attention cannot directly exploit interactions between local audio-visual representations, which limits their expressivity. To address this limitation, we designed a fusion block that attends to local interactions.

\paragraph{Fusion with Dense Audio-Visual Interactions}
We begin by describing a fusion module that attends to dense local interactions. 
Interactions between audio and visual data occur between local regions in the image and the audio spectrogram. As such, we propose to attend to local interactions between audio and visual tokens.
Given a set of $n_a$ audio tokens $\mX^a$ and $n_v$ visual tokens $\mX^v$, aggregation of local multi-modal interactions requires attending to representations from all possible $n_a\times n_v$ pairs, to identify the ones that are most useful for the task at hand.
While we experimented with several ways to represent multi-modal interactions, from bilinear aggregation to kernel-based aggregation, we found that linear aggregation was sufficient to capture the interactions between audio and visual tokens, $\mX^a_i$ and $\mX^v_j$
\begin{eqnarray}
    \mX^{av}_{ij}&=&W^a\mX^a_i+W^v\mX^v_j\\
     \mX^{av}&=&\left[\mX^{av}_{ij}; \forall i=1,\ldots,n_a \forall j=1,\ldots,n_v\right].
\end{eqnarray}
Multi-modal fusion tokens at layer $l$ are then updated by cross-attention over the representation of all local interactions $\mX^{av}$
\begin{equation}
    \mX^{mm}_l = \textit{CrossAttention}\left(\mX^{mm}_{l-1}; \mX^{av}_{l-1}\right)
\end{equation}
where the cross-attention block is computed as follows
\begin{eqnarray}
    \mZ^{mm}_l &=&\mX^{mm}_{l-1} +  \textit{Softmax}\left(\mQ^{mm}_l {\mK^{av}_l}^T\right) \cdot \mV^{av}_l\\
    \mX^{mm}_l &=&\mZ^{mm}_l + \textit{MLP}_l\left(\mZ^{mm}_l\right),
\end{eqnarray}
where $\mK^{av}_l=K_l(\mX^{av}_{l-1})$ and $\mV^{av}_l=V_l(\mX^{av}_{l-1})$ are the key-value pairs for the interaction tokens $\mX^{av}_{l-1}$,  $\mQ^{mm}_l=Q_l({\mX^{mm}_{l-1}})$ the query representations of the fusion tokens $\mX^{mm}_{l-1}$, and $\textit{MLP}_l(\cdot)$ a multi-layer perceptron.

\paragraph{Factorized Audio-Visual Interactions}
While the procedure above can attend to local interactions, it requires the model to explicitly represent all possible $n_a\times n_v$ pairs. Given that the number of audio and visual tokens is typically large, the number of interactions can quickly become intractable, resulting in high memory consumption and low throughput. To address this limitation, we introduce modality-specific aggregation tokens updated by aggregation blocks, $\Phi^{\bar{a}}$ and $\Phi^{\bar{v}}$. The goal of the aggregation blocks is to summarize the large number of audio tokens $\mX^a$ into a small set of $n_{\bar{a}}$ audio aggregation tokens $\bar{\mX}^a$ and the large number of visual tokens $\mX^v$ into a small set of $n_{\bar{v}}$ visual aggregation tokens $\bar{\mX}^v$, respectively. Specifically, at layer $l$, we update the aggregation tokens by cross-attention
\begin{eqnarray}
    \bar{\mX}^a_l &=& \textit{CrossAttention}\left(\bar{\mX}^a_{l-1}, \mX^{a}_{l-1}\right)\\
    \bar{\mX}^v_l &=& \textit{CrossAttention}\left(\bar{\mX}^v_{l-1}, \mX^{v}_{l-1}\right).
\end{eqnarray}

Then, instead of explicitly representing all interactions between audio and visual tokens, we restrict audio-visual interactions to the sets of aggregated tokens.
Formally, factorized interactions between aggregated tokens $\bar{\mX}^a_i$ and $\bar{\mX}^v_j$ are represented as
\begin{eqnarray}
    \bar{\mX}^{av}_{ij}&=&W^{\bar{a}}\bar{\mX}^a_i+W^{\bar{v}}\bar{\mX}^v_j \\
    \bar{\mX}^{av}&=&\left[\bar{\mX}^{av}_{ij}; \forall i=1,\ldots,n_{\bar{a}} \forall j=1,\ldots,n_{\bar{v}}\right]
\end{eqnarray}
and the fusion tokens are updated at layer $l$ by
\begin{equation}
    \mX^{mm}_l = \textit{CrossAttention}\left(\mX^{mm}_{l-1}, \bar{\mX}^{av}_l\right).
\end{equation}
Since we only represent audio-visual interactions between the aggregated uni-modal tokens, the number of possible interactions can be greatly reduced. We show that a reduction by a factor of 700 still maintains the benefits obtained with local dense interactions.

\subsection{Learning Early-Fusion Transformers Through Audio-Visual Masked Auto-Encoding}
Masked auto-encoders have been shown to learn effective representations for a wide variety of input signals, including images~\cite{he2021masked,bao2022beit} and audio~\cite{huang2022amae}.
Formally, the input signal $x$ is partitioned into a set of patch tokens $x=\{x_t\}_{t=1}^T$, which is randomly split into the visible set $X_\vis=\{x_t\}_{t\in\gV}$ and the masked set $X_\mask=\{x_t\}_{t\in\gM}$ ($\gV$ and $\gM$ are non-overlapping index sets). In the case of images, visible and masked tokens, $X_\vis$ and $X_\mask$, are obtained from a grid of small image patches~\cite{he2021masked,bao2022beit}, while in the case of audio signals, tokens are obtained either from small windows of a raw waveform or from small 2D patches of the audio spectrogram~\cite{huang2022amae}.
Then, to learn representations of the input data, a transformer encoder $f$ is first used to encode the visible tokens $\mX_\vis=f(X_\vis)\in\Re^{|| \times d}$, where $d$ denotes the dimensionality of the token representation. These representations together with a series of mask tokens $\mM_\mask$ are then fed to a decoder neural network $g$ that seeks to reconstruct the masked input tokens $\hat{X}_\mask=g(\mX_\vis, \mM_\mask)$. To accomplish that, the whole model is trained to minimize the average $\ell$-2 distance between predicted tokens $\hat{X}_\mask$ and ground-truth tokens $X_\mask$.
\begin{equation}
    \label{eq:mae}
    \gL_{\text{MAE}}\left(X_\mask, \hat{X}_\mask\right) = 
    \frac{1}{|\gM|} \sum_{\substack{x\in X_\mask\\ \hat{x}\in \hat{X}_\mask}} 
    \left\|x - \hat{x}\right\|_2^2.
\end{equation}

This approach can be easily extended to multi-modal signals, like audio-visual data, by simultaneously applying masked reconstruction to both modalities. Specifically, given a paired audio-visual sample $(x^v,x^a)$, we mask both input modalities and encode the visible tokens using our audio-visual early-fusion encoder $f_{mm}$
\begin{equation}
\mX_\vis^v, \mX_\vis^a, \mX_\vis^{mm}=f_{mm}(X_\vis^v, X_\vis^a)
\end{equation}
which, in addition to the uni-modal representations, $\mX_\vis^v$ and $\mX_\vis^a$, also returns a set of multi-modal representations $\mX_\vis^{mm}$ obtained from the learnable fusion tokens. 
These representations are then fed to modality-specific decoders to reconstruct both audio and visual masked tokens, $\hat{X}_\mask^v=g_v(\mX_\vis^{mm}, \mM_\mask^v)$ and $\hat{X}_\mask^a=g_a(\mX_\vis^{mm}, \mM_\mask^a)$, by minimizing the loss
\begin{equation}
    \label{eq:avmae}
    \gL_{\text{AV-MAE}} = \gL_{\text{MAE}}\left(X_\mask^v, \hat{X}_\mask^v\right) + \gL_{\text{MAE}}\left(X_\mask^a, \hat{X}_\mask^a\right)
\end{equation}
where $\gL_{\text{MAE}}$ is the auto-encoder loss of \cref{eq:mae}. 

While multi-modal masked reconstruction has been previously explored in~\cite{huang2022mavil,gong2023contrastive}, these approaches focus on learning separate audio and image encoders, $f_a$ and $f_v$, with multi-modal representations obtained through a late fusion module $f_{mm}$. In contrast, we propose to learn a single multi-modal early fusion encoder $f_{mm}$, and demonstrate that masked reconstruction is especially useful for \textit{learning multi-modal representations} capable of representing \textit{fine audio-visual interactions} through \textit{early fusion} of \textit{local interactions} between audio and visual tokens.

\section{Experiments}
We now evaluate the proposed \method on a variety of audio-visual downstream tasks that require a deep understanding of audio-visual interactions.
The proposed early fusion encoder is first pre-trained using the masked reconstruction objective of~\cref{eq:avmae}. After pertaining, we finetune the resulting model on four downstream tasks, namely audio-visual recognition, and visually-guided sound source separation, localization, and segmentation.

\subsection{Implementation details}
For pre-training, we experiment with two datasets, namely VGG-Sounds~\cite{chen2020vggsound} (with 144k training samples) and AudioSet~\cite{Gemmeke2017audioset} (with 1.73M training samples), both of them containing diverse videos focused on audio events and sound sources.
The audio is represented by log mel spectrograms extracted from $3s$ of audio at a sample rate of $16000$Hz. To compute the log spectrograms, we apply an STFT using approximately 50ms windows with a hop size of 15ms, resulting in an input tensor of size $128 \times 196$ ($128$ mel frequency bands over $196$ time steps).
Since we use image models for the visual component, we randomly extract single frames from within the time window defined by the $3s$ audio snippets. 
Input frames are augmented by random crops with a minimum area of 0.5 and random horizontal flips, and resized into a $224 \times 224$ resolution.

Unless otherwise specified, we initialize the uni-modal transformer blocks using modality-specific encoders pre-trained with uni-modal masked reconstruction objectives~\cite{he2021masked,huang2022amae}. Specifically, we use the ViT-Base model from~\cite{he2021masked} pre-trained on ImageNet~\cite{imagenet_cvpr09}, and the Spec-MAE model (with the same ViT-Base architecture) pre-trained on AudioSet~\cite{Gemmeke2017audioset}.
In our base configuration, we add multi-modal fusion blocks on \textit{all} 12 layers of the model, together with 16 multi-modal fusion tokens, 8 visual aggregation tokens, and 8 audio aggregation tokens. For efficiency, we reduced the dimensionality of the MLP embeddings using an MLP ratio of 1 (instead of the standard 4), as well as the dimensionality of space in which the similarity for self and cross-attention is computed to 16 (as opposed to the standard 64). 

The models were trained using the Adam optimizer~\cite{kingma2014adam}. Ablation and parametric studies were conducted on VGGSound, training for 200 epochs with a total batch size of $512$ (split between 2 GPUs and 4 iterations of gradient accumulation). Large-scale training on VGGSound was conducted for 800 epochs with a total batch size of $1024$ (split between 8 GPUs and 2 iterations of gradient accumulation). Large-scale training on AudioSet was conducted for 200 epochs with a total batch size of $4096$ (split between 8 GPUs and 8 iterations of gradient accumulation). In all cases, we use a base learning rate of $1.5e-4$ adjusted for the effective batch size by the linear scaling rule $lr=blr*bs/256$~\cite{he2021masked}, a 40 epoch learning rate warm-up schedule followed by cosine learning rate decay, and a weight decay of $0.05$.

\subsection{Downstream tasks.}
We fine-tune our model on a variety of downstream tasks.
\begin{description}
\item[Visually-guided sound source separation.] Following~\cite{zhao2018the}, we use a mix-and-separate strategy for training. The pre-trained model is used to obtain audio-visual representations that are fed to a UNet decoder to generate the separated audio. For simplicity, we use a similar decoder architecture to that of~\cite{zhao2018the} (\ie, separation by frequency masking) and train the model using the mean squared error (MSE) loss between the separated audio and the ground truth audio.
    We evaluate on three different datasets: the MUSIC~\cite{zhao2018the} dataset and two subsets from VGG-Sounds~\cite{chen2020vggsound}. MUSIC~\cite{zhao2018the} consists of 448 untrimmed YouTube music videos of solos and duets from 11 instrument categories, where we use 358 solo videos for training and 90 solo videos for evaluation. This dataset was slightly smaller than the original MUSIC dataset since some videos are no longer publicly available to be downloaded. We also use two subsets of VGG-Sounds~\cite{chen2020vggsound}: the VGGSound-Instruments subset~\cite{hu2022mix} which includes 32k video clips of 10s lengths from 36 musical instrument classes, and a broader newly defined "VGGSound-Music" subset which includes 40,908 video clips from 49 music categories for training and 1201 clips for testing. As commonly done in the source separation literature, we measure performance using Signal-to-Distortion Ratio (SDR), Signal-to-Artifact Ratio (SAR), and Signal-to-Interference Ratio (SIR).
    \item[Audio-visual segmentation.] 
    We assess our pre-trained models by fine-tuning them on AVSBench~\cite{zhou2022avs} for the tasks of Single Source Source Segmentation (S4) and Audio-Visual Segmentation with Semantics (AVSS). Task S4 aims to predict pixel-level segmentation maps for the visible sound sources, while in AVSS the model is further required to recognize the class of the sound source (in addition to the segmentation maps). We use the same train/val/test split as in~\cite{zhou2022avs} and measure performance using mIoU and F1 scores.
    \item[Recognition through Linear Probing or Finetuning.] We further evaluate the pre-trained representation on multi-modal recognition tasks by conducting linear probing and fine-tuning evaluations on the VGGSound-Music, VGGSound-All, and the balanced AudioSet dataset. In both cases, we simply attach three linear classifiers on the average pooled visual, audio, and fusion representations, respectively. The three classification logits are then averaged to obtain the final prediction. As usual, on VGGSound datasets, we measure performance by class accuracy, while on AudioSet (a multi-label dataset) we use the average precision (AP) and Area under the ROC curve (AUC) scores averaged across classes.
\end{description}

\begin{table}[t]
\centering
\caption{{\bf Sound source separation} performance on the MUSIC and VGGSound datasets.}
\label{tab: exp_sota_sep}
\resizebox{\linewidth}{!}{\begin{tabular}{l|ccc|ccc|ccc}
    \toprule
    \bf \multirow{2}{*}{Method} & \multicolumn{3}{c|}{\bf MUSIC} & \multicolumn{3}{c|}{\bf VGGS-Instruments}  & \multicolumn{3}{c}{\bf VGGS-Music}  \\
    & SDR & SIR & SAR & SDR & SIR & SAR & SDR & SIR & SAR\\ 	
    \midrule
    NMF~\cite{Virtanen2007monaural} & -0.89 & 2.38	& 6.28 & -3.52 & 0.78 & 6.95 & -5.06 & 0.15 & 7.03 \\
    RPCA~\cite{huang2012rpca}       & 0.78 & 4.62	& 7.58 & 0.36 & 2.38 & 7.95 & -1.68 & 1.57 & 8.26 \\
    SoP~\cite{zhao2018the}          & 3.62 & 7.95 & 9.63 & 2.72 & 5.67 & 9.85	& 1.56 & 4.59 & 10.15 \\
    MP-Net~\cite{xu2019mpnet}       & 3.98 & 8.36 & 9.86 & 3.05 & 6.12 & 10.17	& 1.95 & 5.02 & 10.59 \\
    CCoL~\cite{tian2021cyclic}      & 5.27 & 8.75 & 10.52 & 4.07 & 6.93 & 10.85	& 2.73 & 5.86 & 11.03 \\
    OneAVM~\cite{mo2023oneavm}      & 6.27 & 10.68 & 12.35 & 5.89 & 7.85 & 11.23 & 3.67 & 6.53 & 11.85 \\
    \rowcolor{blue!10}
    \method                         & \bf 7.95 & \bf 12.41 & \bf 12.80 & \bf 6.95 & \bf 9.52 & \bf 13.23 & \bf 5.79 & \bf 8.24 & \bf 13.82 \\
    \bottomrule
\end{tabular}}
\end{table}

\begin{table}
\centering
\caption{{\bf Audio-visual segmentation} performance on the AVSBench dataset for the Sound Source Segmentation (S4) and Audio-Visual Segmentation with Semantics (AVSS) tasks.}
\label{tab: exp_sota_segm}
\scalebox{0.8}{
\begin{tabular}{l|cc|cc}
    \toprule
    \bf \multirow{2}{*}{Method} & \multicolumn{2}{c|}{\bf AVSBench-S4} & \multicolumn{2}{c}{\bf AVSBench-AVSS} \\
    & mIoU & F1 & mIoU & F1 \\	
    \midrule
    AVS~\cite{zhou2022avs,zhou2023avss}	& 78.74 & 87.90 & 29.77 & 35.2 \\
    LAVISH~\cite{lin2023vision} & 80.10 & -- & -- & -- \\
    MMVAE~\cite{mao2023multimodal} & 81.74 & 90.10 & -- & -- \\
    \rowcolor{blue!10}
    \method & \bf 89.94 & \bf 92.34 & \bf 52.05 & \bf 58.29 \\
    \bottomrule
\end{tabular}}
\end{table}

\subsection{Comparison to prior work}
In this work, we propose a novel early fusion architecture with factorized audio-visual interactions trained using a masked auto-encoder framework.
To demonstrate the effectiveness of the proposed method, we extensively compare it to prior work on audio-visual downstream tasks that require a detailed understanding of audio-visual data. Unless otherwise specified, we use our model trained on VGGSound for comparisons to prior work, as this enables us to make the fairest comparisons with the largest number of works that require pre-training.

\paragraph{Sound source separation.} 
Table~\ref{tab: exp_sota_sep} shows the comparison between \method and several prior works, including traditional signal processing methods, NMF~\cite{Virtanen2007monaural} and RPCA~\cite{huang2012rpca}, deep learning methods specialized on the source separation task, Sound-of-Pixels~\cite{zhao2018the} and MP-Net~\cite{xu2019mpnet}, and deep learning methods trained for simultaneous localization and separation, CCoL~\cite{tian2021cyclic} and OneAVM~\cite{mo2023oneavm}. Furthermore, we highlight that while all listed deep learning approaches to source separation use (supervised) ImageNet pre-trained visual encoders, our model is the first to show strong performance using completely unsupervised pre-training through masked reconstruction.
Table~\ref{tab: exp_sota_sep} shows that \method outperforms prior works on all metrics. For example, on the challenging VGGSound-Music, we outperform the recently proposed OneAVM~\cite{mo2023oneavm} by 2.12 SDR and 1.71 SIR, which uses supervised pre-training for the vision component and uses source separation into its second stage pre-training objective.
Although the different models use different architectures for their audio-visual encoders (which make direct comparison challenging to analyze), the improvements demonstrated on this task indicate that visual source separation can benefit from early fusion to extract rich representations from fine audio-visual interactions.

\paragraph{Audio-Visual Segmentation.}
We also validate the effectiveness of \method for audio-visual segmentation on the AVSBench dataset~\cite{zhou2022avs,zhou2023avss}. We compared the proposed approach to recent state-of-the-art methods (AVS~\cite{zhou2022avs}, LAVISH~\cite{lin2023vision} and MMVAE~\cite{mao2023multimodal}) on the S4 and AVSS downstream tasks (sound source segmentation without and with semantics, respectively). As is shown in~\cref{tab: exp_sota_segm}, \method outperforms all prior work on both tasks, demonstrating the effectiveness of the proposed early fusion architecture for audio-visual segmentation.

\paragraph{Audio-visual classification.} 
While the above experiments demonstrate the effectiveness of \method on audio-visual tasks that require a detailed understanding of audio-visual interactions, we also evaluated the learned representations on recognition tasks.
In particular, we compared our model to uni-modal masked reconstruction methods, MAE~\cite{he2021masked} and AudioMAE~\cite{huang2022amae}.
We also compared to a late-fusion audio-visual masked reconstruction model, Audio-Visual MAE, which uses the same audio-visual reconstruction objective of~\cref{eq:avmae} but does not leverage an early fusion encoder, and the CAV-MAE model~\cite{gong2023contrastive} which uses a contrastive learning objective in addition to masked reconstruction. For a fair comparison, we use the same ViT-Base architecture for the uni-modal encoders and tune the model on image-audio pairs obtained from the VGG-Sound dataset.

As can be seen in \cref{tab: exp_sota_cls}, we outperform both uni-modal pre-trained encoders on linear probing and fine-tuning by significant margins.
\method outperforms the image MAE~\cite{he2021masked} and AudioMAE~\cite{huang2022amae} by 19.62\% and 13.26\% on VGGSound-Music and 19.08\% and 12.49\% on the full VGGSound dataset.
These direct comparisons to uni-modal encoders highlight the importance of learning joint audio-visual representations for multi-modal recognition.
Furthermore, our \method also significantly outperforms Audio-Visual MAE which optimizes that same audio and visual reconstruction objective of~\cref{eq:avmae} but does not leverage an early fusion encoder.

It should be noted that multi-modal recognition could be further improved by learning from audio-video data (instead of audio-image pairs).
However, video modeling is beyond the scope of this work. The improvements in \cref{tab: exp_sota_cls} demonstrate nevertheless that even tasks like audio-visual classification of sound sources can benefit from more deeply integrated multi-modal representations.

\begin{table}[t]
\centering
\caption{{\bf Audio-visual classification} performance on VGGSound-Music and VGGSound-All datasets, using linear probing (LP) and finetuning (FT) protocols. Performance is measured in terms of classification accuracy [\%].
}
\label{tab: exp_sota_cls}
\scalebox{0.8}{
\begin{tabular}{l|cc|cc}
    \toprule
    \multirow{2}{*}{\bf Method} & \multicolumn{2}{c|}{\bf VGGS-Music} & \multicolumn{2}{c}{\bf VGGS-All}  \\
    & LP & FT & LP & FT  \\ 	
    \midrule
    (Image) MAE~\cite{he2021masked}  & 48.25 & 53.18 & 37.12 & 43.86 \\
    AudioMAE~\cite{huang2022amae} & 52.73 & 58.29 & 42.05 & 49.53 \\
    CAV-MAE~\cite{gong2023contrastive} & 59.18 & 67.58 & 49.83 & 55.32 \\
    \rowcolor{blue!10}
    \method & \bf 65.32	& \bf 72.25 & \bf 53.08 & \bf 58.19 \\
    \bottomrule
\end{tabular}}
\end{table}

\subsection{Understanding \method}
The previous section demonstrated the efficacy of early fusion of local audio-visual interactions, with the proposed method outperforming the state-of-the-art on a wide range of audio-visual tasks. In this section, we conduct a thorough analysis of our framework, revealing emergent properties in our model, and assessing the benefits of uni-modal pre-training, fusion depth, and other major design choices of the proposed methodology. To provide a full-picture comparison, we evaluate all models on five downstream tasks, namely linear probing and fine-tuning on VGGSound, sound source separation on VGGSound-Music, sound source localization on Flickr-SoundNet, and audio-visual segmentation with semantics on AVSBench.

\begin{table}[t]
\centering
\caption{Ablation study of major components of \method. 
\label{tab: ab_property}}
\resizebox{0.95\linewidth}{!}{
\begin{tabular}{ccccccccc}
    \toprule
    \bf (\#) &
    \bf \thead{Unimodal\\Pre-Training} & 
    \bf \thead{Fusion} & 
    \bf \thead{Local AV\\Interactions} &
    \bf \thead{Linear\\Acc} & 
    \bf \thead{Sep\\SDR} & 
    \bf \thead{Segm\\mIoU} \\
    \midrule
    \rowcolor{blue!10}
    (1) & \cmark & Early & \cmark 
    & \bf 53.08 & \bf 6.53 & \bf 52.05 \\
    (2) & \textcolor{ForestGreen}{\xmark} & Early & \cmark 
    & 44.36 & 2.03 & 29.55 \\
    (3) & \cmark & Early & \textcolor{ForestGreen}{\xmark} 
    & 51.19 & 6.15 & 48.71 \\
    (4) & \cmark & \textcolor{ForestGreen}{Mid}  & \cmark 
    & 51.21 & 5.85 & 48.04 \\
    (5) & \cmark & \textcolor{ForestGreen}{Late} & \cmark 
    & 46.52 & 5.32 & 44.32 \\
    (6) & \cmark & \textcolor{ForestGreen}{None} & \textcolor{ForestGreen}{\xmark} 
    & 39.67 & 4.23 & 38.32 \\
    \bottomrule
\end{tabular}}
\end{table}

\paragraph{Ablation studies.}
We begin our analysis with a series of ablation studies in Table~\ref{tab: ab_property} to measure the impact of three important components of the proposed framework, namely 1) uni-modal pre-training, 2) early fusion, and 3) audio-visual interactions. All models use the same backbone architecture for the uni-modal encoders and are trained to optimize the Audio-Visual MAE objective.
Model \#1 is the full \method framework, and the remaining models relax each of the three components in turn.
Model \#2 skips uni-modal pre-training, significantly hurting performance when compared to model \#1 on all tasks, emphasizing the value of uni-modal pre-training, when learning deeply integrated audio-visual encoders. 
Model \#3 performs early fusion but uses the "token fusion" strategy similar to that used in MBT~\cite{nagrani2022attention}. Token fusion attends to uni-modal representations and cannot directly capture audio-visual interactions, limiting the expressivity of the fusion block, and the overall performance of the model.
Models \#4, \#5 and \#6 use the same fusion modules with dense interactions, but perform mid-fusion (at layers 9-12), late-fusion (at layer 12), and no fusion at all. The results indicate that mid-fusion outperforms, late-fusion which in turn is substantially better than no-fusion. Nevertheless, early fusion achieves the best results across all downstream tasks.

\begin{table}[t]
\centering
\caption{Exploration studies on the efficiency of factorized interactions.}
\label{tab: ab_efficiency}
\resizebox{0.95\linewidth}{!}{
\begin{tabular}{ccccccc}
    \toprule
    \bf \thead{Early\\Fusion}  & 
    \bf \thead{Factorized\\Interactions} & 
    \bf \thead{Speed\\(video/sec)} & 
    \bf \thead{Memory\\(GB)} &   
    \bf \thead{Linear\\Acc} & 
    \bf \thead{Sep\\SDR} & 
    \bf \thead{Segm\\mIoU} \\
    \midrule
    \xmark  & \xmark & \bf 14.5 & \bf 10.3 & 46.52 & 5.32 & 44.32 \\
    \cmark  & \xmark & 4.3 & 31.6  & 52.82 & 6.30 & \bf 52.13 \\
    \rowcolor{blue!10}
    \cmark  & \cmark & 12.8 & 13.1 & \bf 53.08 & \bf 6.53 & 52.05 \\
    \bottomrule
\end{tabular}}
\end{table}

\paragraph{Efficiency of factorized interactions.}
While audio-visual interactions are crucial for the performance of early fusion, they also introduce a significant computational overhead, if not handled properly. Table~\ref{tab: ab_efficiency} lists the VRAM consumption and model throughput during inference for three models: a baseline model without fusion, an early fusion model with dense interactions, and our proposed early-fusion model with factorized interactions.
Dense interactions enhance the model's effectiveness on subsequent tasks, but at a high computational cost, both in terms of throughput (3.3 times slower) and memory usage (3.1 times higher). Our proposed factorized interactions significantly cut down the computational overhead, while maintaining (and even slightly enhancing) the performance of early fusion with dense interactions.

\begin{figure}[t]
\centering
\includegraphics[width=0.95\linewidth]{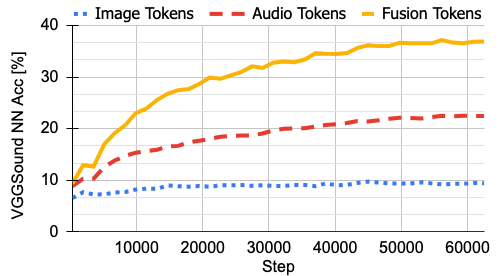}
\caption{Nearest neighbor accuracy (NNAcc) vs training steps (Step) using Audio, Image, Fusion Tokens from an \method trained on VGGSounds.}
\label{fig: vis_nearest_neighbor}
\end{figure}

\paragraph{Emergent semantics.}
In addition to robust performance across various tasks, the emergent semantics in fusion tokens is worth highlighting. 
Despite the masked auto-encoder objective typically favoring representations rich in low-level details, fusion tokens surprisingly yield higher-level semantic representations.
\cref{fig: vis_nearest_neighbor} shows the nearest neighbor retrieval performance of our \method using either the learned visual, audio, or fusion tokens as queries. As can be seen, the retrieval using fusion tokens is substantially better than uni-modal representations (\ie better aligned with the semantics of the dataset). This suggests that fusion tokens aggregate information that is indicative of high-level semantics, while uni-modal representations encode the low-level details required for masked reconstruction.

\newcommand{\fusionlayer}{
\begin{tabular}{cccccc}
    \toprule
    \bf \# & 
    \bf \thead{Linear\\Acc} & 
    \bf \thead{Sep\\SDR} & 
    \bf \thead{Segm\\mIoU} \\
    \midrule
    \rowcolor{blue!10}
    12 & \bf 53.08 & \bf 6.53 & \bf 52.05 \\
    9  & 52.76 & 6.15 & 49.93 \\
    6  & 52.29 & 6.12 & 49.02 \\
    3  & 51.21 & 5.85 & 48.04 \\
    1  & 46.52 & 5.32 & 44.32 \\
    0  & 39.67 & 4.23 & 38.32 \\
    \bottomrule
\end{tabular}
}

\newcommand{\fusiontkns}{
\begin{tabular}{ccccccc}
    \toprule
    \bf \# &
    \bf \thead{Linear\\Acc} & 
    \bf \thead{Sep\\SDR} & 
    \bf \thead{Segm\\mIoU} \\
    \midrule
    1 & 51.86 & 5.94 & 50.67 \\
    8 & 51.97 & 5.83 & 51.14 \\
    \rowcolor{blue!10}
    16 & \bf 53.08 & \bf 6.53 & \bf 52.05 \\
    32 & 52.18 & 6.43 & 50.60 \\
    \bottomrule
\end{tabular}
}

\newcommand{\aggtkns}{
\begin{tabular}{ccccccccc}
    \toprule
    \bf \# &
    \bf \thead{Linear\\Acc} & 
    \bf \thead{Sep\\SDR} & 
    \bf \thead{Segm\\mIoU} \\
    \midrule 
    4  & 51.78 & 5.68 & 49.59 \\
    \rowcolor{blue!10}
    8  & \bf 53.08 & \bf 6.53 & \bf 52.05 \\
    16 & 52.25 & 6.16 & 50.89 \\
    \bottomrule
\end{tabular}
}

\begin{table}[!tb]
    \centering
    \caption{Impact of early fusion and fusion tokens.
    \label{tab: ab_robust}}
    \begin{subtable}{0.43\linewidth}
        \centering
        \resizebox{\linewidth}{!}{\fusionlayer}
        \caption{\footnotesize \# Fusion layers}
       \label{tab: ab_layer}
    \end{subtable}
    \quad
    \begin{subtable}{0.43\linewidth}
        \centering
        \resizebox{\linewidth}{!}{\fusiontkns}
        \caption{\footnotesize \# Fusion tokens}
        \label{tab: ab_num_fusion_token}
    \end{subtable}
    \begin{subtable}{0.43\linewidth}
        \centering
        \resizebox{\linewidth}{!}{\aggtkns}
        \caption{\footnotesize \# Aggr tokens}
        \label{tab: ab_num_agg_token}
    \end{subtable}
\end{table}

\paragraph{Impact of miscellaneous design choices.}
Lastly, we examine the effects of the number of fusion layers, fusion tokens and aggregation tokens in \cref{tab: ab_robust}.
\cref{tab: ab_layer} reveals that early fusion is crucial for optimal performance, with model performance improving as fusion depth increases.
Furthermore, \method can benefit from a relatively large number of tokens, with the performance saturating after 16 fusion tokens and 8 aggregation tokens.

\section{Conclusion}

In this work, we present \method, a simple yet effective early fusion approach with dense interactions that achieves efficient audio-visual pre-training for audio-visual masked auto-encoders. 
We introduce learnable fusion tokens to aggregate modality-specific information with early fusion from each transformer block of audio and visual encoders, where two variants of parallel and sequential flows are proposed to achieve early fusion between fusion tokens and audio-visual patch tokens. 
Furthermore, we leverage multi-modal blocks with dense interactions to fuse fusion tokens and patches across audio-visual transformer blocks and achieve efficient downstream fine-tuning with factorized attention blocks. 
Empirical experiments on Flick-SoundNet, VGG-Instruments, VGG-Music, VGGSound-All, and AVSBench datasets demonstrate the state-of-the-art performance of our \method in linear probing, fine-tuning classification, visual sound localization, sound separation, and audio-visual segmentation.

{
    \small
    \bibliographystyle{style-cvpr/ieeenat_fullname}
    \bibliography{reference}
}

\newpage

\appendix
\section*{Appendix}

In this supplementary material, we first provide detailed implementation details on each audio-visual main downstream task and additional analysis of \method.
We also provide a few demonstrations of sound source localization, segmentation, and separation.

\section{Implementation Details}

\subsection{Audio-visual Classification}
We conduct audio-visual classification on VGGSounds using two different downstream evaluation protocols: linear probing and fine-tuning.
In both cases, we attach a linear layer to the pre-trained encoder as a classification head. Specifically, we (average) pool audio representations from all time-frequency patches into a single global audio representation, visual representations from all spatial patches into a global image representation, and the representations obtained for the fusion tokens into a global audio-visual representation. The input to the linear classifier is a subset of these three global features. In the main paper, we used only audio and visual representations. Below, we provide an analysis using different subsets that include fusion tokens.
Linear probe evaluations only train the linear head to evaluate the quality of pre-trained features, while fine-tuning evaluations finetune the full model to assess \method's ability to provide strong initializations. In both cases, the models are trained for 50 epochs using the Adam optimizer~\cite{kingma2014adam} with a learning rate of $1e-4$ and batch size of $128$.

\subsection{Visually-Guided Sound Source Localization and Segmentation}
We further evaluate our models on sound source localization and segmentation, using a supervised transfer learning protocol. 
We assess source localization on Flickr-SoundNet, following the work of~\cite{Senocak2018learning}. For \method, the localization prediction maps $\hat{y}$ are defined as the cosine similarity between the global audio representation and the local visual representations. The full model is then trained to minimize the average per-pixel binary cross entropy to ground truth maps $y$ using the Adam optimizer~\cite{kingma2014adam} for $30$ epochs, with a learning rate of $1e-4$ and a batch size of $128$. To compare with prior work, we use the same downstream evaluation protocol (both the prediction head and training procedure), while using the different objectives proposed in the original papers to pre-train the same ViT-Base backbones (as used by \method).
Source segmentation is evaluated on AVSBench using the training protocol of prior work~\cite{zhou2022avs}. Specifically, to compute the segmentation maps, we fuse the global audio representation with the local visual representations, and upsample these localized audio-visual representations using an upsampling decoder with three sequential up-convolution blocks. The full model is then trained to minimize the binary cross entropy (BCE) between the segmentation maps predicted from the upsampling decoder and ground-truth masks.
The model is trained for 20 epochs using the Adam optimizer~\cite{kingma2014adam} with a learning rate of $1e-4$ and a batch size of $128$. 

\subsection{Sound Source Separation} 
For sound source separation, we follow the training protocol of~\cite{zhao2018the,mo2023oneavm}. We attach an audio U-Net decoder to the pre-trained audio-visual encoders.
The decoder receives the representations from an audio mixture and from the target image (containing the object whose sound should be separated).
Then, through a series of transposed convolutions and an output head, the decoder predicts a time-frequency separation mask, which is used to modulate the STFT of the input mixture in order to predict the separated sound source.
Similarly to~\cite{zhao2018the,mo2023oneavm}, the model is trained to minimize the binary cross-entropy between the predicted separation masks and binary target masks, which identify the time-frequency bins where the target source is the most dominant component in the mixture.
The model is trained for 20 epochs using the Adam optimizer~\cite{kingma2014adam} with a learning rate of $1e-4$ and a batch size of $128$.

\begin{table*}[!htp]
\renewcommand\tabcolsep{10.0pt}
\renewcommand{\arraystretch}{1.1}
\centering
\caption{The impact of initializing \method with models pre-trained for uni-modal masked reconstruction at scale.}
\label{tab: ab_pretrain}
\scalebox{0.8}{
\begin{tabular}{ccccccccc}
    \toprule
    \bf \thead{Unimodal\\Pre-Train} & 
    \bf \thead{Early\\Fusion} & 
    \bf \thead{AV\\Interactions} &
    \bf \thead{Dense\\Interactions} &
    \bf \thead{Linear\\Acc} & 
    \bf \thead{Finetune\\Acc} & 
    \bf \thead{Loc\\Prec} & 
    \bf \thead{Sep\\SDR} & 
    \bf \thead{Segm\\mIoU} \\
    \midrule
    \xmark & \cmark & \cmark & \cmark & 40.65 & 52.79 & 57.16 & 4.53 & 35.05 \\
    \cmark  & \cmark & \cmark & \cmark & \bf 42.75	& \bf 55.28 & \bf 58.78 & \bf 4.83	& \bf 36.41  \\
    \bottomrule
\end{tabular}}
\end{table*}

\section{Impact of \method initialization}
The unimodal components of \method are initialized with pre-trained weights from MAE~\cite{he2021masked} and AudioMAE~\cite{huang2022amae}. To further assess the impact of this initialization, we've trained our model when trained from scratch on the VGG-Sounds dataset.
Table~\ref{tab: ab_pretrain} compares the two models (trained with and without unimodal pre-trained initialization) on a variety of downstream tasks.
We observe that \method performs quite well, even without the strong initialization. For example, \method without unimodal initialization significantly outperforms the pre-trained model without early fusion (Table 4, row 2 of main paper). Nevertheless, the strong initialization can still benefit the learned representations, even when training on a relatively large-scale dataset like VGG-Sounds.

\section{Demonstrations of model capabilities}

\subsection{Sound Source Localization and Segmentation}

\begin{figure*}[t]
\centering
\includegraphics[width=0.95\linewidth]{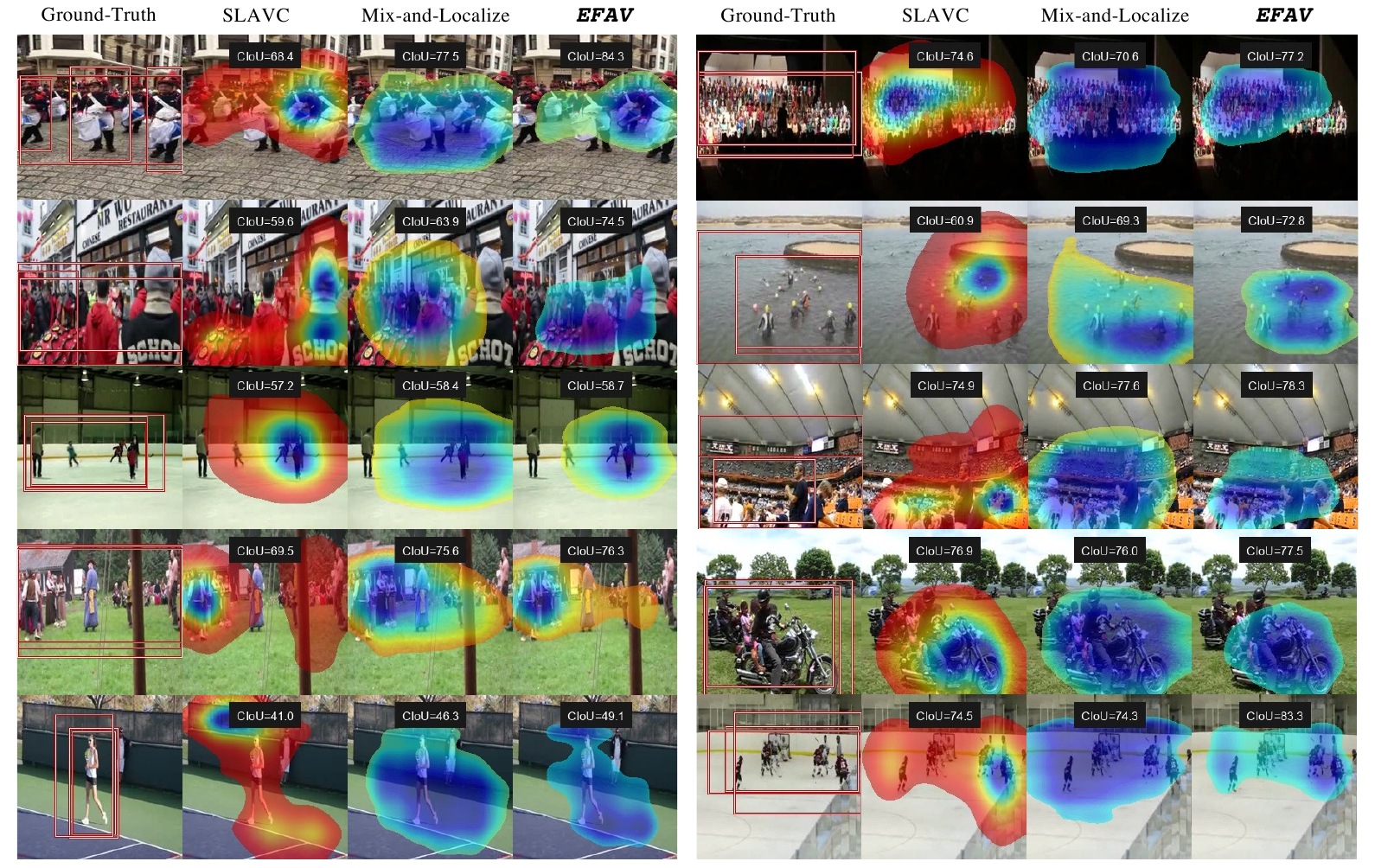}
\caption{ Qualitative visualization of sound source localization. The proposed \method produces more accurate and high-quality localization maps for each source.}
\label{fig: vis_loc}
\end{figure*}

\begin{figure*}[t]
\centering
\includegraphics[width=0.95\linewidth]{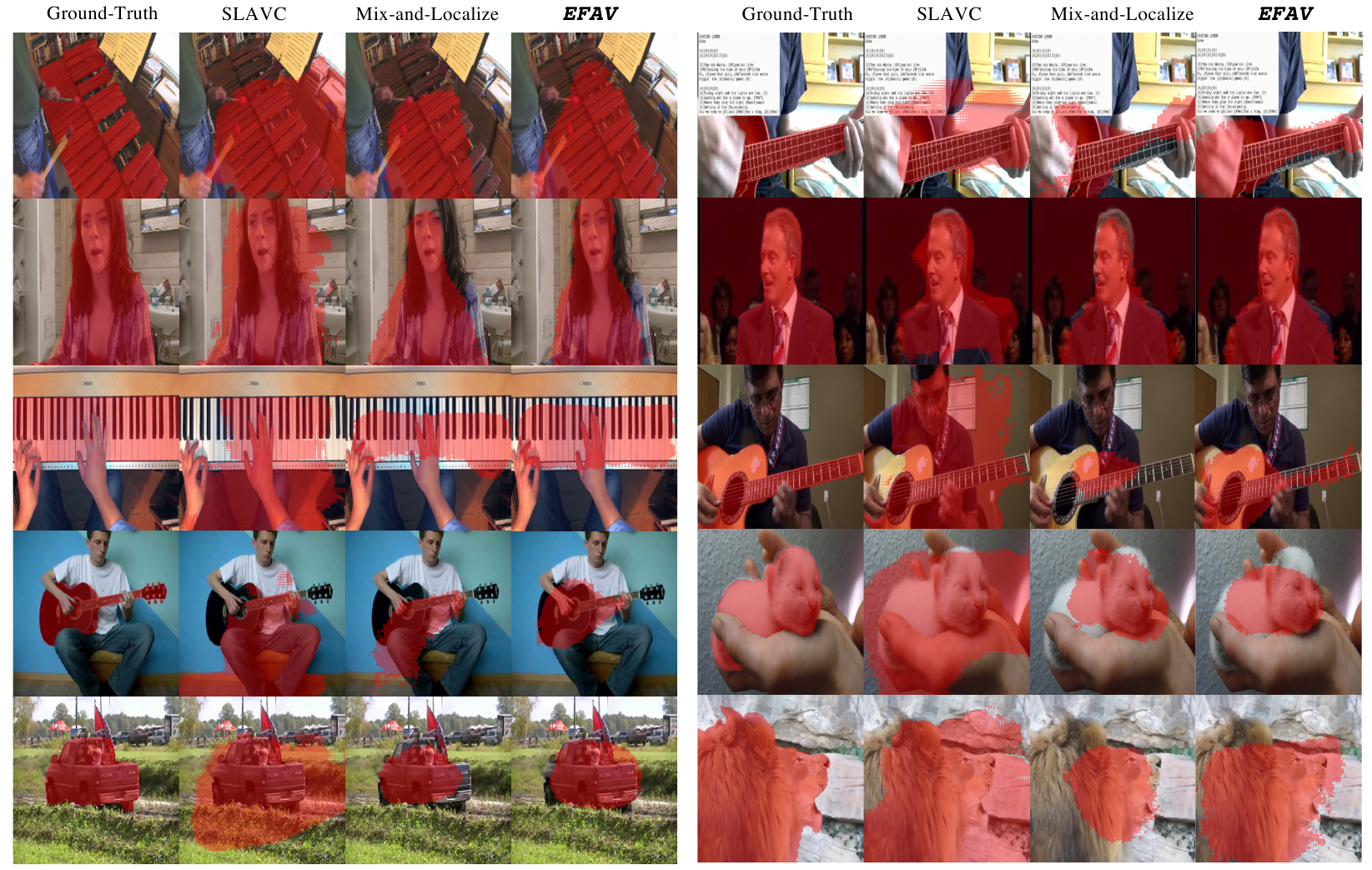}
\caption{ Qualitative visualization of sound source segmentation. The proposed \method produces more accurate and high-quality segmentation masks for each source.}
\label{fig: vis_seg}
\end{figure*}

In order to qualitatively evaluate the localization and segmentation masks, we compare the proposed \method with SLAVC~\cite{mo2022SLAVC} and Mix-and-Localize~\cite{hu2022mix} on sound source localization and segmentation in Figure~\ref{fig: vis_loc} and Figure~\ref{fig: vis_seg}.
We can observe that the quality of localization maps and masks generated by our framework are indeed superior to prior state-of-the-art methods. For example, the performance gains of \method over Mix-and-Localize~\cite{hu2022mix} (a strong multi-task approach with both localization and separation) can be easily seen in these demonstrations.
These visualizations showcase the effectiveness of the proposed \method in sound source localization and segmentation.

\subsection{Sound Source Separation} 
\begin{figure*}[t]
\centering
\includegraphics[width=0.95\linewidth]{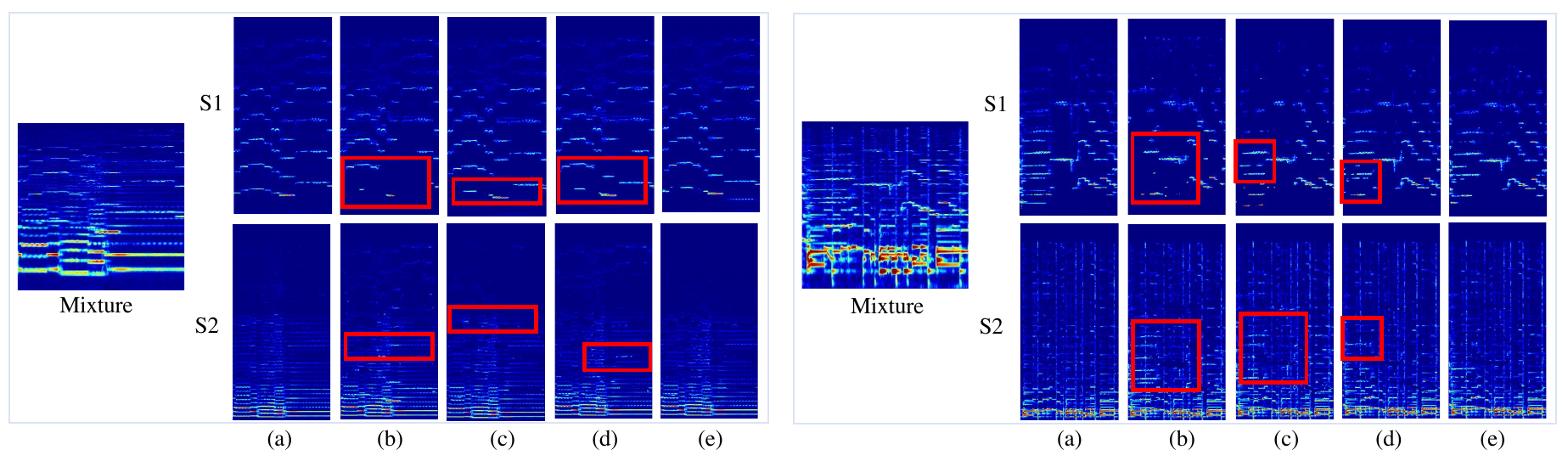}
\caption{Qualitative visualization of sound source separation. (a) Ground-Truth; (b) Sound-of-Pixels; (c) CCoL; (d) OneAVM; (e) \method (ours).
The proposed \method separates each source more accurately.}
\label{fig: vis_sep}
\end{figure*}

In Figure~\ref{fig: vis_sep}, we qualitatively compare the proposed \method with audio-visual separation baselines, Sound-of-Pixels~\cite{zhao2018the}, CCoL~\cite{tian2021cyclic}, and OneAVM~\cite{mo2023oneavm} in terms of reconstructed source spectrograms. 
Once again, we can observe the higher quality of spectrograms generated by our method, even when compared to OneAVM~\cite{mo2023oneavm}, the state-of-the-art model which jointly optimized for recognition, localization, and source separation. These visualizations further showcase the superiority of the proposed \method in sound source separation.

\section{Potential Negative Impacts} 

The proposed method utilizes a masked auto-encoder framework to enable efficient training of audio-visual early fusion architecture to improve multi-modal recognition, sound source localization, segmentation, and source separation from user-uploaded web videos, which might cause the model to learn internal biases in the data. 
For example, the model could fail to localize, segment, and separate certain rare but critical sound sources in the videos. 
These bias issues should be carefully addressed when it comes to the deployment of real applications.

\end{document}


\maketitle

\input{APPENDIX/manuscript}

{
    \small
    \bibliographystyle{ieeenat_fullname}
    \bibliography{reference}
}
